\documentclass{article}

\usepackage{subfigure}

\usepackage{hyperref}

\usepackage[accepted]{icml2021}

\usepackage[utf8]{inputenc} 
\usepackage[T1]{fontenc}    
\usepackage{url}            
\usepackage{booktabs}       
\usepackage{amsfonts}       
\usepackage{nicefrac}       
\usepackage{microtype}      
\usepackage{amsthm}
\usepackage{mathtools, amssymb, algorithm, bm, float,caption, graphicx, color}
\theoremstyle{definition}

\newtheorem{theorem}{Theorem}
\newtheorem{definition}{Definition}
\newtheorem{lemma}{Lemma}
\newtheorem{remark}{Remark}
\newtheorem{corollary}{Corollary}

\newcommand{\iu}{\mathrm{i}\mkern1mu}

\renewcommand{\tilde}{\widetilde}

\def\beq{\begin{equation}}
\def\eeq{\end{equation}}
\def\beqa{\begin{eqnarray}}
\def\eeqa{\end{eqnarray}}
\def\beqan{\begin{eqnarray*}}
\def\eeqan{\end{eqnarray*}}

\def\Z{{\mathbb{Z}}}

\DeclareMathOperator*{\unif}{unif}
\DeclareMathOperator*{\argmin}{arg\,min}

\DeclareMathOperator{\dom}{dom}
\DeclareMathOperator{\vect}{vec}

\newcommand\independent{\protect\mathpalette{\protect\independenT}{\perp}}
\def\independenT#1#2{\mathrel{\rlap{$#1#2$}\mkern2mu{#1#2}}}

\def\Exp{\mathbb{E}}
\def\var{\mathrm{var}}

\def\PL2{\stackrel{PL(2)}{=}}

\newcommand{\inv}{^{-1}}
\newcommand{\zero}{\mathbf{0}}

\newcommand{\fbf}{\mathbf{f}}

\newcommand{\Kbf}{\mathbf{K}}
\newcommand{\kbf}{\mathbf{k}}

\newcommand{\xbf}{\mathbf{x}}

\newcommand{\xbfhat}{\wh{\mathbf{x}}}

\newcommand{\ybf}{\mathbf{y}}

\newcommand{\zbf}{\mathbf{z}}

\newcommand{\Abf}{\mathbf{A}}

\newcommand{\Cbf}{\mathbf{C}}

\newcommand{\Ibf}{\mathbf{I}}

\newcommand{\Xbf}{\mathbf{X}}
\newcommand{\Xbfhat}{\widehat{\mathbf{X}}}
\newcommand{\Xbfthat}{\widehat{\tilde{\mathbf{X}}}}
\newcommand{\Ybf}{\mathbf{Y}}
\newcommand{\Zbf}{\mathbf{Z}}

\newcommand{\inner}[1]{\langle{#1}\rangle}

\def\mubf{{\boldsymbol \mu}}

\def\Gammabf{{\boldsymbol \Gamma}}

\newcommand{\omegabf}{\bm{\omega}}

\def\etabf{{\boldsymbol \eta}}

\def\xibf{{\boldsymbol \xi}}
\def\Xibf{{\boldsymbol \Xi}}

\newcommand{\MAP}{_{\text{\sf MAP}}}

\newcommand{\tran}{^{\text{\sf T}}}
\newcommand{\herm}{^{\text{\sf H}}}

\def\eqd{\stackrel{d}{=}}

\def\PLp{\stackrel{PL(p)}{=}}
\def\Norm{{\mathcal N}}
\def \CNorm{\mathcal{C}\mathcal{N}}

\def \kbft{\tilde{\mathbf{k}}}
\def \Kbft{\tilde{\mathbf{K}}}

\def \Xbft{\tilde{\mathbf{X}}}

\def \Xt{\tilde{X}}

\def\Ybft{\tilde{\mathbf{Y}}}

\def \Xibft{\tilde{\mathbf{\Xi}}}

\RequirePackage{xcolor,bm,setspace}
\RequirePackage{mathrsfs}

\providecommand{\old}[1]{ }

\providecommand{\mest}{\text{\sf m-est}}
\providecommand{\ridge}{\text{\sf ridge}}

\providecommand{\wh}{\widehat}
\newcommand{\norm}[1]{\left\|#1\right\|}
\newcommand{\tnorm}[1]{\left\|#1\right\|_2}
\newcommand{\fnorm}[1]{\left\|#1\right\|_{\rm F}}
\newcommand{\fnormsm}[1]{\|#1\|_{\rm F}}

\providecommand{\Real}{\mathbb{R}}
\providecommand{\Complex}{\mathbb{C}}

\providecommand{\g}{\bm{g}}

\providecommand{\Z}{\bm{Z}}

\providecommand{\fbf}{\mathbf{f}}

\providecommand{\kbf}{\mathbf{k}}

\providecommand{\xbf}{\mathbf{x}}
\providecommand{\ybf}{\mathbf{y}}
\providecommand{\zbf}{\mathbf{z}}
\providecommand{\Abf}{\mathbf{A}}

\providecommand{\Cbf}{\mathbf{C}}

\providecommand{\Fbf}{\mathbf{F}}

\providecommand{\Ibf}{\mathbf{I}}

\providecommand{\Kbf}{\mathbf{K}}

\providecommand{\Xbf}{\mathbf{X}}
\providecommand{\Ybf}{\mathbf{Y}}
\providecommand{\Zbf}{\mathbf{Z}}

\providecommand{\mcA}{\mathcal{A}}
\providecommand{\mcB}{\mathcal{B}}
\providecommand{\mcC}{\mathcal{C}}

\providecommand{\mcF}{\mathcal{F}}

\providecommand{\mcL}{\mathcal{L}}

\providecommand{\mcN}{\mathcal{N}}

\providecommand{\mcR}{\mathcal{R}}

\providecommand{\mcW}{\mathcal{W}}
\providecommand{\mcX}{\mathcal{X}}

\newcommand{\indep}{\rotatebox[origin=c]{90}{$\models$}}

\begin{document}

\twocolumn[
\icmltitle{Asymptotics of Ridge Regression in Convolutional Models}



\begin{icmlauthorlist}
\icmlauthor{Mojtaba Sahraee-Ardakan}{ECE,STATS}
\icmlauthor{Tung Mai}{Adobe}
\icmlauthor{Anup Rao}{Adobe}
\icmlauthor{Ryan Rossi}{Adobe}
\icmlauthor{Sundeep Rangan}{NYU}
\icmlauthor{Alyson K.~Fletcher}{ECE,STATS}
\end{icmlauthorlist}
\icmlaffiliation{ECE}{ECE, UCLA}
\icmlaffiliation{STATS}{STATS, UCLA}
\icmlaffiliation{NYU}{ECE, NYU}
\icmlaffiliation{Adobe}{Adobe Research}
\icmlcorrespondingauthor{Mojtaba Sahraee-Ardakan}{msahraee@ucla.edu}
\icmlkeywords{Convolutional Models, Ridge Regression, Asymptotic Error}
\vspace{0.3in}]
\printAffiliationsAndNotice{}
\begin{abstract}
Understanding generalization and estimation error of estimators for simple models such as linear and generalized linear models has attracted a lot of attention recently. This is in part due to an interesting observation made in machine learning community that highly over-parameterized neural networks achieve zero training error, and yet they are able to generalize well over the test samples. This phenomenon is captured by the so called double descent curve, where the generalization error starts decreasing again after the interpolation threshold. A series of recent works tried to explain such phenomenon for simple models. In this work, we analyze the asymptotics of estimation error in ridge estimators for convolutional linear models. These convolutional inverse problems, also known as deconvolution, naturally arise in different fields such as seismology, imaging, and acoustics among others. Our results hold for a large class of input distributions that include i.i.d.\ features as a special case. We derive exact formulae for estimation error of ridge estimators that hold in a certain high-dimensional regime. We show the double descent phenomenon in our experiments for convolutional models and show that our theoretical results match the experiments.
\end{abstract}
\section{Introduction}\label{sec:intro}
Increasingly powerful hardware along with deep learning libraries that efficiently use these computational resources have allowed us to train ever larger neural networks. Modern neural networks are so over-parameterized that they can perfectly fit random noise \cite{zhang2016understanding,li2020gradient}. With enough over-parameterization, they can also achieve zero loss over training data with their parameters moving only slightly away from the initialization \cite{allen2018convergence, soltanolkotabi2018theoretical, du2018gradient, du2018gradient2, li2018learning, zou2020gradient}. Yet, these models generalize well on test data and are widely used in practice \cite{zhang2016understanding}. In fact, some recent work suggest that it is best practice to use as large a model as possible for the tasks in hand \cite{DBLP:journals/corr/abs-1811-06965}. This seems contrary to our classical understanding of generalization where increasing the complexity of the model space to the extent that the training data can be easily interpolated indicates poor generalization. Most statistical approaches explain generalization by controlling some notion of capacity of the hypothesis space, such as VC dimension, Rademacher complexity, or metric entropy \cite{anthony2009neural}. 
Such approaches that do not incorporate the implicit regularization effect of the optimization algorithm fail to explain generalization of over-parameterized deep networks \cite{kalimeris2019sgd,oymak2019overparameterized}.

The so-called \emph{double descent} curve where the test risk starts decreasing again by over-parameterizing neural networks beyond the interpolation threshold is widely known by now \cite{belkin2019reconciling, nakkiran2019deep}. Interestingly, such phenomenon is not unique to neural networks and have been observed even in linear models \cite{dobriban2018high}. Recently, another line of work has also connected neural networks to linear models. In  \cite{jacot2018neural}, the authors show that infinitely wide neural networks trained by gradient descent behave like their linearization with respect to the parameters around their initialization. The problem of training such wide neural networks with square loss then turns into a kernel regression problem in a RKHS associated to a fixed kernel called the \emph{neural tangent kernel} (NTK). The NTK results were later extended to many other architectures such as convolutional networks and recurrent neural networks \cite{li2019enhanced, alemohammad2020recurrent, yang2020tensor}. Trying to understand the generalization in deep networks and explaining such phenomenon as the double descent curve has attracted a lot of attention to theoretical properties of kernel methods as well as simple machine learning models. Such models, despite their simplicity, can help us gain a better understanding of machine learning models and algorithms that might be hard to achieve just by looking at deep neural networks due their complex nature.

Double descent has been shown in linear models \cite{dobriban2018high, belkin2019two, hastie2019surprises}, logistic regression \cite{ deng2019model}, support vector machines \cite{montanari2019generalization}, generalized linear models \cite{emami2020generalization}, kernel regression \cite{liang2020just}, random features regression \cite{mei2019generalization, hastie2019surprises}, and random Fourier feature regression \cite{liao2020random} among others. Most of these works consider the problem in a doubly asymptotic regime where both the number of parameters and the number of observations go to infinity at a fixed ratio. This is in contrast to classical statistics where either the number of parameters is assumed to be fixed and the samples go to infinity or vice versa. In practice, the number of parameters and number of samples are usually comparable and therefore the doubly asymptotic regime provides more value about the performance of different models and algorithms. In this work we study the performance of ridge estimators in convolutional linear inverse problems in this asymptotic regime. Unlike ordinary linear inverse problems, theoretical properties of the estimation problem in convolutional models has not been studied, despite their wide use in practice, e.g.\ in solving inverse problems with deep generative priors \cite{ulyanov2018deep}. 

Beyond machine learning, inverse problems involving convolutional measurement models are often called deconvolution and are encountered in many different fields. In astronomy, deconvolution is used for example to deblur, sharpen, and correct for optical aberrations in imaging \cite{starck2002deconvolution}. In seismology, deconvolution is used to separate seismic traces into a source wavelet and an impulse response that corresponds to the layered structure of the earth \cite{treitel1982linear, mueller1985source}. In imaging, it is used to correct for blurs caused by the point spread function, sharpen out of focus areas in 3D microscopy \cite{mcnally1999three}, and to separate neuronal spikes for calcium traces in calcium imaging \cite{friedrich2017fast} among others. In practical applications, the convolution kernel might not be known and should either be estimated from the physics of the problem or jointly with the unknown signal using the data.

\paragraph{Summary of Contributions.} We analyze the performance of ridge estimator for convolutional models in the proportional asymptotics regime. Our main result (Theorem \ref{thm:main}) characterizes the limiting joint distribution of the true signal and its ridge estimate in terms of the spectral properties of the data. As a result of this theorem, we can provide an exact formula to compute the mean squared error of ridge estimator in the form of a scalar integral (Corollary \ref{cor:estimation_error}). Our assumptions on the data are fairly general and include many random processes as an example as opposed to i.i.d. features only. Even though our theoretical results hold only in a certain high dimensional limit, our experiments show that its prediction matches the observed error even for problems of moderate size. We show that our result can predict the double descent curve of the estimation error as we change the ratio of number of measurements to unknowns.

\paragraph{Prior Work.} Asymptotic error of ridge regression for ordinary linear inverse problems (as opposed to convolutional linear inverse problem considered in this work) is studied in \cite{dicker2016ridge} for isotropic features. Asymptotics of ridge regression for correlated features is studied in \cite{dobriban2018high}. Error of ridgeless (minimum $\ell_2$-norm interpolant) regression for data generated from a linear or nonlinear model is obtained in \cite{hastie2019surprises}. These works use results from random matrix theory to derive closed form formulae for the estimation or generalization error. For features with general i.i.d. prior other than Gaussian distribution, approximate message passing (AMP) \cite{donoho2010message, bayati2011dynamics} or vector approximate message passing (VAMP) \cite{rangan2019vector} can be used to obtain asymptotics of different types of error. Instead of a closed form formula, these works show that the asymptotic error can be obtained via a recursive equation that is called the \emph{state evolution}. In \cite{deng2019model}, the authors use convex Gaussian min-max theorem to characterize the performance of maximum likelihood as well as SVM classifiers with i.i.d. Gaussian covariates. In \cite{emami2020generalization}, the problem of learning generalized linear models is reduced to an inference problem in deep networks and the results of \cite{pandit2020inference} are used to obtain the generalization error.

\paragraph{Notation.} We use uppercase boldface letters for matrices and tensors, and lowercase boldface letters for vectors. For a matrix $\Abf$, its $i$th row and column is denoted by $\Abf_{i*}$ and $\Abf_{*i}$ respectively. A similar notation is used to show slices of tensors. The submatrix formed by columns $i$ through $j-1$ of $\Abf$ is shown by $\Abf_{*, i:j}$. Standard inner product for vectors, matrices, and tensors is represented by $\inner{\cdot, \cdot}$. $\Norm(0, 1)$ and $\CNorm(0, 1)$ denote standard normal and complex normal distributions respectively. Finally, $[n]= \{1,\dots, n\}$.
\section{Problem Formulation}
We consider the inverse problem of estimating $\Xbf$ from $\Ybf$ in the convolutional model
\begin{equation}
    \Ybf = \Kbf * \Xbf + \Xibf,\label{eq:conv_model}
\end{equation}
where $\Xbf\in \Complex^{n_x\times T}$, $\Ybf\in \Complex^{n_y\times T}$, $\Kbf\in \Complex^{n_y\times n_x \times k}$ with $\Kbf_{i**}\in \Complex^{n_x\times k}$ being the $i$th convolutional kernel of width $k$, and $\Xibf$ is a noise matrix with the same shape as $\Ybf$ and i.i.d.\ zero-mean complex normal entries $\CNorm(0, \sigma^2)$. See Appendix \ref{app:complex_normal_dist} for a brief overview of complex normal distribution. The circular convolution in equation \eqref{eq:conv_model} is defined as
\begin{align}
    \Ybf_{i*} &= \Kbf_{i**} * \Xbf + \Xibf_{i*} \label{eq:conv_def_per_channel}\\
    \Ybf_{it} &= \inner{\Kbf_{i**}, \Xbf_{*,t:t+k}} + \Xibf_{it}, \quad i\in [n_y], t\in [T]\label{eq:conv_def}
\end{align}
Note that in \eqref{eq:conv_def} we are not using the correct definition of the convolution operation where the kernel (or the signal $\Xbf$) is reflected along the time axis, but rather we are using the common definition used in machine learning.

We consider the inference problem in the Bayesian setting where the signal $\Xbf$ is assumed to have a prior that admits a density (with respect to Lebesgue measure) $p(\Xbf)$. Further, we assume that rows of $\Xbf_i$ are i.i.d.\ such that this density factorizes as
\begin{equation}
    p(\Xbf) = \prod_{i=1}^{n_x} p(\Xbf_{i*}).\label{eq:prior_factorization}
\end{equation}
The convolution kernel $\Kbf$ is assumed to be known with i.i.d.\ entries drawn from $\CNorm(0, 1/(n_yk))$. Given this statistical model, the posterior is
\begin{equation}
    p(\Xbf|\Ybf)\propto p(\Xbf, \Ybf) = p(\Xbf)p(\Ybf|\Xbf),\label{eq:posterior_factorization}
\end{equation}
where with some abuse of notation, we are using $p(\cdot)$ to represent the densities of all random variables to simplify the notation. From the Gaussianity assumption on noise we have
\begin{equation}
   \Ybf_{i*}|\Xbf \sim \CNorm(\Kbf_{i**} * \Xbf, \sigma^2 \Ibf),\label{eq:likelihood_gaussian_noise}
\end{equation}
where $\Ibf$ is the identity matrix of size $T\times T$.

Given the model in \eqref{eq:posterior_factorization}, one can consider different types of estimators for $\Xbf$. Of particular interest are regularized M-estimators
\begin{equation}
    \Xbfhat_\mest = \argmin_{\Xbf} \mcL(\Xbf, \Ybf) + \mcR(\Xbf),\label{eq:m-est}
\end{equation}
where $\mcL$ is a loss function and $\mcR$ corresponds to the regularization term. Taking negative log-likelihood as the loss and negative log-prior as the regularization we get the \emph{maximum a posteriori} (MAP) estimator
\begin{equation}
    \Xbfhat\MAP := \argmin_{\Xbf} - \log p(\Ybf|\Xbf) -\log p(\Xbf),\label{eq:MAP}
\end{equation}
which selects mode of the posterior as the estimate.

In this work we are interested in analyzing the performance of ridge-regularized least squares estimator which is another special case of the regularized M-estimator in \eqref{eq:m-est} with square loss and $\ell_2$-norm regularization
\begin{equation}
    \Xbfhat_\ridge = \argmin_{\Xbf} \fnorm{\Ybf - \Kbf*\Xbf}^2 + \lambda\fnorm{\Xbf}^2,\label{eq:ridge_est}
\end{equation}
where $\lambda$ is the regularization parameter. By Gaussianity of the noise, this can also be thought of as $\ell_2$-regularized maximum likelihood estimtor.
Given an estimate $\Xbfhat$, one is usually interested in performance of the estimator based on some metric. In Bayesian setting, the metric is usually an average risk (averaged over the prior and the randomness of data)
\begin{equation}
    R = \Exp \ell(\Xbf, \Xbfhat),
\end{equation}
where $\ell$ is some loss function between the true parameters and the estimates. The most widely used loss is the squared error where $\ell(\Xbf, \Xbfhat) = \|\Xbf-\Xbfhat\|_{\rm F}^2$. Our theoretical results exactly characterize the mean squared error (MSE) of ridge estimator \eqref{eq:ridge_est} in a certain high-dimensional regime described below.
\section{Main Result}\label{sec:main-result}
Similar to other works in this area \cite{bayati2011dynamics,rangan2019vector, pandit2020inference}, our goal is to analyze the average case performance of the ridge estimator in \eqref{eq:ridge_est} for the convolutional model in \eqref{eq:conv_model} in a certain high dimensional regime that is called \emph{large system limit} (LSL).

\subsection{Large System Limit}\label{sec:large_system_limit}
We consider a sequence of problems indexed by $T$ and $n_x$. We assume that $k:=k(T)$ and $n_y:=n_y(n_x)$ are functions of $T$ and $n_x$ respectively and
\begin{equation}
    \lim_{n_x\rightarrow \infty}\frac{n_y(n_x)}{n_x}=\delta\in(0,\infty),\quad \lim_{T\rightarrow \infty} \frac{k(T)}{T}=\beta\in(0,1].\nonumber
\end{equation}
This doubly asymptotic regime where both the number of parameters and unknowns are going to infinity at a fixed ratio is sometimes called proportional asymptotic regime in the literature. We assume that the entries of the convolution kernel $\Kbf$ and the noise $\Xibf$ converge empirically with second order to random variables $K$ and $\Xi$, with distributions $\CNorm(0, \sigma_K^2/(kn_y))$ and $\CNorm(0, \sigma^2)$ respectively. See Appendix \ref{app:empirical_conv_definitions} for definition of empirical convergence of random variables. 

\subsubsection{Assumptions on $\Xbf_{i*}$}\label{sec:random_process_assumptions}
Next, based on \cite{peligrad2010central}, we state the distributional assumptions on the rows of $\Xbf$ that we require for our theory to hold. Let $\{\xi_{t}\}_{t\in \Z}$ be a stationary ergodic Markov chain defined on a probability space $(S, \mcF,P)$ and let $\pi(\cdot)$ be the distribution of $\xi_0$. Note that stationarity implies all $\xi_t$s have the same marginal distribution. Define the space of functions
\begin{equation}
    \mcL_0^2(\pi) := \left\{h|\Exp_\pi[h(\xi_0)]=0, \Exp_\pi[h^2(\xi_0)] < \infty\right\}.
\end{equation}
Define $\mcF_k:=\sigma(\{\xi_t\}_{t\leq k})$, the $\sigma$-algebra generated by $\xi_t$ up to time $k$ and let $x_t = h(\xi_t)$ for some $h\in \mcL_0^2(\pi)$. We assume that the process $\{x_t\}_{t\in\Z}$ satisfies the regularity condition
\begin{equation}
    \Exp[x_0|\mcF_{-\infty}] = 0, \quad P-\text{almost surely}.
\end{equation}
The class of processes that satisfy these conditions is quite large and includes i.i.d. random processes as an example. It also includes causal functions of i.i.d. random variables of the form $X_n = f(\xi_k, k\leq n)$ where $\xi_k$ is i.i.d. such as autoregressive (AR) processes and many Markov chains. See \cite{peligrad2010central} for more examples satisfying these conditions.

We assume that each row $\Xbf_{i*}$ of $\Xbf$, is an i.i.d. sample of a process that satisfies the conditions mentioned. Let $\Xbft_i(\omega)$ be its (normalized) Fourier transform (defined in \eqref{eq:Fourier_transform}). and define $g(\omega):=\lim_{T\rightarrow \infty}\Exp|\Xbft_i(\omega)|^2$. As shown in \cite{peligrad2010central}, since the rows are i.i.d., this limit is the same for all the rows. Also, $g(\omega)$ is proportional to the spectral density of the process that generates the rows of $\Xbf$
\begin{equation}
    g(\omega) = \frac{1}{2\pi}\sum_{t=-\infty}^\infty c_n \exp(-\iu\omega t), \quad c_t = \Exp[\Xbf_{i0}\Xbf_{it}].\label{eq:spectral_density}
\end{equation}
As we will see in the next section, $g(\omega)$ plays a key role in characterization of estimation error of ridge estimator in convolutional linear inverse problems that have such processes as inputs.

\subsection{Asymptotics of Ridge Estimator}
The main result of our paper characterizes the limiting distribution to which the the Fourier transform of the true signal $\Xbft_0(\omega)$ and Fourier transform of the estimated signal $\Xbfthat_\ridge(\omega)$ converge. The proof can be found in Section \ref{sec:proof}. In the following $\mcB$ and $\mu$ represent Borel $\sigma$-algebra and Lebesgue measure respectively.
\begin{theorem}\label{thm:main}
Under the assumptions in Section \ref{sec:large_system_limit}, as $n_x, n_y, T, k \rightarrow \infty$, over the product space $([0, 2\pi]\times S^{n_x}, \mcB\times \mcF^{n_x}, \mu\times P^{n_x})$ the Ridge estimator satisfies
    \begin{equation}
         \left[\begin{matrix}
        \Xbft_0(\omega)\\ \Xbfthat_\ridge(\omega)
        \end{matrix}\right] \stackrel{d, PL(2)}{=} \left[\begin{matrix}
        \sqrt{g(U)}Z_0\\ \alpha(\sqrt{g(U)}Z_0 + \tau(g(U)) Z_1)
        \end{matrix}\right],\nonumber
    \end{equation}
    where $U\sim\unif([0, 2\pi]), Z_0, Z_1\sim \mcC\mcN(0, 1)$ where $\mcC\mcN(0,1)$ is the standard complex normal distribution, $U, Z_0$ and $Z_1$ are independent of each other, $\alpha$ is the smaller root of the quadratic equation
\begin{equation}
     \lambda = \frac{(1-\alpha)(1- \alpha/\delta)}{\alpha},\label{eq:lambda_alpha_eq_thm}
\end{equation}
and 
\begin{equation}
    \tau^2(g(U)) = \frac{\sigma^2 +(1-\alpha^2)g(U)/\delta}{1 - {\alpha^2}/{\delta}}.\label{eq:tau_for_ridge_thm}
\end{equation}
\end{theorem}
The convergence in this theorem is weak convergence for $\omega$ and PL(2) for $\Xbft_0(\omega)$ and $\Xbfthat_\ridge(\omega)$. This convergence result allows us to find the asymptotic mean squared error of ridge estimator for the convolutional model as an integral.
\begin{corollary}\label{cor:estimation_error}
Under the same assumptions as in Theorem \ref{thm:main}, ridge estimator satisfies
\begin{multline}
    \lim_{n_x\rightarrow\infty}\lim_{T\rightarrow\infty}\frac{1}{n_x T}\fnormsm{\Xbfthat_\ridge-\Xbft}^2 \\= \int_0^{2\pi} \left((\alpha - 1)^2 g(\omega) + \alpha^2 \tau(g(\omega))\right)d\omega.\label{eq:error_integral_form}
\end{multline}
\end{corollary}
\begin{remark}
As shown in the proof Lemma \ref{lem:asymptotics_of_ridge}, for $\lambda \geq 0$, the quadratic equation \eqref{eq:lambda_alpha_eq_thm} always has two real positive solutions the smaller of which determines the error.
\end{remark}
\begin{remark}
The $1/\sqrt{T}$ scaling in our definition of Fourier operator in Section \ref{sec:main_technical_lemmas} makes it a unitary operator, i.e.\ $\ell_2$ norm is preserved under the Fourier transform and its inverse. This implies
\begin{equation}
    \|\Xbfhat_\ridge-\Xbf\|_{\rm F}= \|\Xbfthat_\ridge-\Xbft\|_{\rm F}.
\end{equation}
Therefore, the result of Corollary \ref{cor:estimation_error} also holds in time domain.  
\end{remark}
\begin{remark} \label{rem:double_descent_iid} If rows $\Xbf_{i*}$ have zero mean i.i.d. entries, then the correlation coefficients $c_t$ in \eqref{eq:spectral_density} are all zero except for $c_0$. Therefore, $g(\omega) = g$ where $g$ is constant. Hence, in this case, the integrand in \eqref{eq:error_integral_form} would be a constant and the estimation error across all the frequencies would be the same as the estimation error in the ordinary ridge regression as in Lemma \ref{lem:asymptotics_of_ridge}, i.e.\ the error vs.\ $\delta$ would be exactly the same as the double descent curve in ordinary ridge regression with i.i.d.\ priors. In other words, the double descent curve for ordinary ridge regression carries over to the convolutional ridge regression for i.i.d.\ priors (see Figure \ref{fig:iid_Gaussian}).

\end{remark}
\section{Proof}\label{sec:proof}
In this section we present the proof of Theorem \ref{thm:main}. Before presenting the details of the proof, it is helpful to see an overview of the proof. 

\subsection{Proof overview}
We first show that convolutional models turn into ordinary linear models for each frequency in Fourier domain. We then show that ridge estimators in time domain can also be written as ridge estimators in frequency domain. This uses the fact that Fourier transform matrix, with appropriate normalization is a unitary matrix, and $\ell_2$ norm is preserved under unitary transformation, i.e.\ it is an isometry. Next we use properties of Fourier transform of random processes to show that under certain conditions, they asymptotically converge to a Gaussian process in frequency domain that is independent across different frequencies for almost every frequency. These together allow us to turn the ridge estimation in time domain into multiple ridge estimators in frequency domain, one for each frequency. We then use theoretical properties of ridge estimators to derive estimation error for each of these ridge estimators and integrate them over frequencies to derive our main result.

Our proof is based on previous results for asymptotic error of ridge estimators for ordinary linear inverse problems. This has been studied in many works \cite{dicker2016ridge, dobriban2018high, hastie2019surprises} where the authors take advantage of the fact that ridge estimators have a closed form solution that can be analyzed, e.g.\ using results from random matrix theory. In this work we use approximate message passing \cite{donoho2010message, bayati2011dynamics} to derive the asymptotic error of ridge estimators.

\subsection{Main Technical Lemmas}\label{sec:main_technical_lemmas}
In order to prove Theorem \ref{thm:main}, we need several lemmas. We first characterize the convolutional model in \eqref{eq:conv_model} in Fourier domain. For $\omega\in \{0, 1\cdot \frac{2\pi}{T}, \dots, (T-1)\cdot \frac{2\pi}{T}\}=:\Omega$, let $\Xbft_j(\omega)$ be the discrete (circular) Fourier transform (DFT) of $\Xbf_{j*}$
\begin{equation}
    \Xbft_j(w) = \frac{1}{\sqrt{T}}\sum_{t=0}^{T-1} \Xbf_{jt}\exp(-\iu\omega t).\label{eq:Fourier_transform}
\end{equation}
If we let $\Fbf$ to denote the $T$-point unitary DFT matrix, i.e. $\Fbf_{mn}=1/\sqrt{T}\exp(-\iu 2\pi mn/T)$,  then this equation can be written in matrix form as $\Xbft_{i}(*)=\Xbf_{i*}\Fbf$. Define $\Ybft_i(\omega)$, $\Kbft_{ij}(\omega)$, and $\Xibft(\omega)$ similarly. Note that in these definitions, we have included a $1/\sqrt{T}$ factor which is usually not included in the definition of Fourier transform, but since this makes the Fourier matrix unitary, it eases our notation slightly. With this definition we have $\Fbf\tran = \Fbf$ and $\Fbf^*\Fbf = \Ibf$.

\begin{lemma}[Convolutional models in Fourier domain] \label{lem:convolutional_model_in_Fourier_domain}
The convolutional model in \eqref{eq:conv_model} can be written in Fourier domain as
\begin{equation}
    \Ybft(\omega) = \sqrt{T}\Kbft_{**}(\omega)\Xbft(\omega) + \Xibft(\omega), \quad \forall \omega \in \Omega.
\end{equation}
\end{lemma}
\begin{proof}
Taking Fourier transform of Equation \eqref{eq:conv_def_per_channel} and using the convolution theorem we get
\begin{equation}
    \Ybft_i(\omega) = \sqrt{T}\sum_{j=1}^{n_x} \Kbft_{ij}(\omega)\Xbft_j(\omega) + \Xibft_i(\omega).
\end{equation}
Note that the $\sqrt{T}$ factor on the right hand is due to our definition of Fourier transform where we have used a $1/\sqrt{T}$ factor to make the Fourier operator unitary. Rewriting this equation in matrix form gives us the desired result.
\end{proof}

The next lemma characterizes the ridge estimator in \eqref{eq:ridge_est} in frequency domain.
\begin{lemma}\label{lem:ridge_estimator_in_Fourier}
The ridge estimator in \eqref{eq:ridge_est} in frequency domain is equivalent to solving separate ordinary ridge regressions for each $\omega\in \Omega$:
\begin{multline}
    \Xbfthat_\ridge(\omega) = \argmin_{\Xbft(\omega)} \tnorm{\Ybft(\omega) - \sqrt{T}\Kbft(\omega)\Xbft(\omega)}^2\\ + \lambda\tnorm{\Xbft(\omega)}^2.\label{eq:ridge_est_freq}
\end{multline}
\end{lemma}
\begin{proof}
Since the Fourier matrix is unitary, we have 
\begin{align*}
    \fnorm{\Xbf} &= \fnorm{\Xbf\Fbf} = \fnormsm{\Xbft}\\
    \fnorm{\Ybf-\Kbf*\Xbf} &= \fnorm{(\Ybf-\Kbf*\Xbf)\Fbf} = \fnormsm{\Ybft - \sqrt{T}\Kbft\Xbft}, 
\end{align*}
where $\Kbft\Xbft$ is a tensor-matrix product defined as $(\Kbft\Xbft)(\omega) = \Kbft(\omega)\Xbft(\omega)$. Then, a change of variable $\Xbft = \Xbf\Fbf$ in \eqref{eq:ridge_est} proves the lemma.
\end{proof}

\begin{lemma}\label{lem:kernel_fourier}
If the kernel $\Kbf$ has i.i.d.\ $\CNorm(0,\sigma_K^2/(kn_y))$ entries, then for each $\omega\in \Omega$, $\sqrt{T}\Kbft(\omega)$ has i.i.d. complex normal entries $\mcC\mcN(0, \sigma_K^2/n_y)$.

\end{lemma}
\begin{proof}
The DFT of the kernel is
\begin{equation}
    \Kbft_{ij}(\omega) = \frac{1}{\sqrt{T}}\sum_{t=0}^{k-1} \Kbf_{ijt}\exp(-\iu\omega t).
\end{equation}
This is a linear combination of complex Gaussian random variables and therefore, $\Kbft$ is a tensor with jointly complex Gaussian entries. Clearly, $\Exp(\Kbft) = \zero$ and for $(i,j)\neq (i', j')$, using independence of $\Kbf_{ij*}$ and $\Kbf_{i'j'*}$ we have
\begin{equation}
    \Exp \Kbft_{ij}(\omega)\Kbft_{i'j'}(\omega') = 0, \quad \Exp \Kbft_{ij}(\omega)\Kbft^*_{i'j'}(\omega') =0 \quad \forall \omega, \omega',\nonumber
\end{equation}
which proves that for any $\omega$, $\Kbft(\omega)$ has independent entries and all the dependence in $\Kbft$ is across different frequencies.

It remains to find the variance and relation of each entry of $\Kbft(\omega)$. Let $\kbf:=\Kbf_{ij*}$ for some $i$ and $j$ be a row vector, and let $\Kbft :=\Kbft_{ij}(*)$. Then we have $\kbft = \kbf \Fbf$ and the variance of $\sqrt{T}\kbft_m$ is
\begin{align}
    \gamma &= T\Exp[\kbft_m\kbft_m^*] = T\Exp[\kbf\Fbf_{*m}\Fbf_{*m}^* \kbf\tran]\\
           &= T\Fbf_{m*}\Exp[\kbf\tran\kbf]\Fbf^*_{m*} = \frac{T\sigma_K^2}{kn_y}\Fbf_{m*}
           \left[\begin{matrix}\Ibf_k & \zero\\ \zero & \zero\end{matrix}\right]\Fbf^*_{m*}\\
           &= \frac{\sigma_K^2}{kn_y}\sum_{t=0}^{k-1}\exp(\frac{2\pi\iu tm}{T})\exp(\frac{-2\pi\iu tm}{T})\\
           &= \frac{\sigma_K^2}{n_y}.
\end{align}
Similarly, the relation is
\begin{align}
    c(m) &= T\Exp[\kbft_m\kbft_m\tran] = T\Exp[\kbf\Fbf_{*m}\Fbf_{m*} \kbf\tran]\nonumber\\
           &= T\Fbf_{m*}\Exp[\kbf\tran\kbf]\Fbf_{*m} = \frac{\sigma_K^2}{k}\Fbf_{m*}
           \left[\begin{matrix}\zero & \zero\nonumber\\ \zero & \zero\end{matrix}\right]\Fbf_{*m}\nonumber\\
           &= 0.\nonumber
\end{align}
Therefore, for each $\omega$, $\Kbft_{ij}(\omega)\sim \CNorm(0, \sigma_K^2/n_y)$ and they are i.i.d. for all $i,j$.
\end{proof}
\begin{remark}
Observe that the scaling of variance of entries of $\Kbf$ with $1/k$ is crucial to get a non-trivial distribution for entries of $\sqrt{T}\Kbft(\omega)$ as we take the limit $T\rightarrow \infty$.
\end{remark}
\begin{lemma}\label{lem:noise_fourier}
If noise $\Xibf$ has i.i.d.\ $\CNorm(0,\sigma^2)$ entries, then $\Xibft$ has i.i.d. complex normal entries $\Xibft_{ij}(\omega)\sim \CNorm(0,\sigma^2)$,  i.e.
\begin{equation}
    \Xibf_{ij}(\omega) \eqd \frac{\sigma^2}{2}Z_1 + \frac{\sigma^2}{2}Z_2, \quad Z_1, Z_2 \sim\Norm(0,1), Z_1\independent Z_2.\nonumber
\end{equation}
\end{lemma}
\begin{proof}
    The proof is similar to the proof of Lemma \ref{lem:kernel_fourier} with $k=T$.
\end{proof}
Lemma \ref{lem:noise_fourier} is the complex analogue of the fact that distribution of vectors with i.i.d. Gaussian entries is invariant under orthogonal transformations. 

As stated in Appendix \ref{app:empirical_conv_definitions}, for a Gaussian random sequence, convergence in the first and second moments implies convergence in Wasserstein distance which is equivalent to $PL(2)$ convergence. Therefore, Lemma \ref{lem:kernel_fourier} and \ref{lem:noise_fourier} also imply that the entries of kernel and noise for each frequency are converging empirically with second order to i.i.d. complex Gaussian random variables. As shown in the appendix, this convergence is stronger than convergence in distribution.

Next, we mention a result about Fourier transform of random processes from \cite{peligrad2010central}.

\begin{lemma}[Fourier transform of random processes \cite{peligrad2010central}]\label{lem:Fourier_of_random_process} Let $\{X_t\}_{t\in\Z}$ be a stationary ergodic process that satisfies the assumptions in Section \ref{sec:random_process_assumptions}. Let $\Xt(\omega)$ be its (normalized) Fourier transform and $g(\omega):=\lim_{T\rightarrow \infty}\Exp|\Xt(\omega)|^2$. Then on the product space $([0, 2\pi]\times S, \mcB\times \mcF, \mu\times P)$ we have
\begin{equation}
    \Xt(\omega) \eqd \sqrt{g(U)}\mcC\mcN(0, 1),
\end{equation}
where $U\sim\unif([0, 2\pi])$ is independent of $\mcC\mcN(0,1)$.
\end{lemma}

Lemma \ref{lem:Fourier_of_random_process} allows us to characterize the limiting distribution of each row $\Xbf_{i*}$ of the input in the frequency domain, i.e.\ distribution of $\Xbft_i(\omega)$ as $T\rightarrow \infty$.

All the lemmas so far allow us to look at the convolutional model in the frequency domain. We need one last lemma to characterize the asymptotics of ridge regression in high dimensions.
\begin{lemma}[Asymptotics of ridge regression] \label{lem:asymptotics_of_ridge} Consider the linear model
$\ybf = \Abf \xbf_0 + \xibf$, where $\xbf\in\Real^{n_x}, \xibf\in\Real^{n_y}$, and $\Abf\in \Real^{n_y\times n_x}$ all have i.i.d. components with $\xbf_i\sim\Norm(0, \sigma_x^2)$, $\xibf_i\sim\Norm(0, \sigma^2)$, and $\Abf_{ij}\sim\Norm(0, 1/n_y)$. Then the ridge estimator
\begin{equation}
    \xbfhat_\ridge = \argmin_\xbf \tnorm{\ybf-\Abf\xbf}^2 + \lambda \tnorm{\xbf}^2\label{eq:ridge_real}
\end{equation}
as $n_x, n_y\rightarrow \infty$ with $n_y/n_x\rightarrow\delta$ satisfy
\begin{equation}
    \left[\begin{matrix} \xbf_0\\
    \xbfhat_\ridge
    \end{matrix}\right] \PL2 \left[\begin{matrix} X_0\\
    \alpha(X_0 + \tau(\sigma_x) Z)
    \end{matrix}\right],\label{eq:ridge_asymptotic_AMP}
\end{equation}
where $X_0\sim p_X$, $Z\sim\Norm(0,1)$ independent of $X_0$, $\alpha$ is the smaller root of the quadratic equation
\begin{equation}
     \lambda = \frac{(1-\alpha)(1- \alpha/\delta)}{\alpha},\label{eq:lambda_alpha_eq_proof}
\end{equation}
and 
\begin{equation}
    \tau^2(\sigma_x^2) = \frac{\sigma^2 +(1-\alpha^2)\sigma_x^2/\delta}{1 - {\alpha^2}/{\delta}}.\label{eq:tau_for_ridge}
\end{equation}
Therefore, we almost surely have
\begin{equation}
\lim_{n_x\rightarrow \infty}1/n_x\tnorm{\xbf_\ridge-\xbf_0}^2=(\alpha - 1)^2\sigma_X^2 + \alpha^2 \tau^2(\sigma_x).\nonumber
\end{equation}

\end{lemma}
\begin{proof}
    This is a consequence of using \emph{approximate message passing} (AMP) algorithm to solve \eqref{eq:ridge_real}. See Appendix \ref{app:AMP} for a brief introduction to AMP and Appendix \ref{app:AMP_ridge} for the proof of this lemma. Here we briefly mention the sketch of the proof. AMP is a recursive algorithm that originally was proposed to solve linear inverse problems but since has been applied to many different problems. The key property of AMP algorithms is that a set of equations called the state evolution (SE) exactly characterize the performance of AMP at each iteration. As shown in the appendix, the AMP recursions to solve ridge regression in \eqref{eq:ridge_real} are
\begin{align}
    \xbf^{t+1} &= \alpha(\Abf\tran \zbf^t + \xbf^t),\\
    \zbf^t &= \ybf - \Abf \xbf^t + \frac{\alpha}{\delta}\zbf^{t-1},
\end{align}
where $\alpha$ is the smaller root of the quadratic equation in \eqref{eq:lambda_alpha_eq_proof}. In the appendix, we prove that for $\lambda \geq 0$, the roots of this equation are real and positive, and only for the smaller root the AMP algorithm converges. Finally, using the state evolution, we obtain that the ridge estimator and true values of $\xbf$ jointly converge as in \eqref{eq:ridge_asymptotic_AMP}, and $\tau^2(\sigma_x)$ in \eqref{eq:tau_for_ridge} is the fixed point value of state evolution recursion in \eqref{eq:state_evolution_ridge}.
\end{proof}
This lemma allows us to find asymptotics of ridge regression for real linear inverse problems. Even though the $\tau$ in \eqref{eq:tau_for_ridge} depends also on $\alpha, \delta,$ and noise variance, we have only made the dependence on $\sigma_x$ explicit, as all the other parameters will be fixed for the ridge regression problem for each frequency, but $\sigma_x$ could change as a function of frequency.
\begin{remark}
The exact same result holds for complex valued ridge regression \emph{mutatis mutandis}, i.e. by changing normal distributions $\Norm(0, \cdot)$ to complex normal distributions $\CNorm(0, \cdot)$.
\end{remark}
\begin{remark}
The requirements of Lemma \ref{lem:asymptotics_of_ridge} can be relaxed. Rather than requiring $\xbf, \Abf$, or $\xibf$ to have i.i.d.\ Gaussian entries, we only need them to converge PL(2) to random variables with these distributions. 
\end{remark}
\subsection{Proof of Theorem \ref{thm:main}}
We now have all the ingredients to prove Theorem \ref{thm:main}. Lemma \ref{lem:ridge_estimator_in_Fourier} allows us to find Fourier transform of Ridge estimator in \ref{eq:ridge_est} using a series of ridge regressions in Fourier domain. Lemmas \ref{lem:kernel_fourier} and \ref{lem:noise_fourier}  show that the Fourier transform of the convolution kernel and the noise and the signal asymptotically have complex Gaussian distributions with i.i.d. components for each frequency. Then, Lemma \ref{lem:Fourier_of_random_process} shows that
\begin{equation}
    \Xbft_i(\omega) \eqd \sqrt{g(U)}\mcC\mcN(0, 1),
\end{equation}
where $g(\cdot)$ is defined in the Lemma.

Next, the complex version of Lemma \ref{lem:asymptotics_of_ridge} would give us the asymptotic error of ridge estimator on the product space $([0, 2\pi]\times S, \mcB\times \mcF, \lambda\times P)$ in the limit
\begin{equation}
    \left[\begin{matrix} \Xbft_0(\omega)\\
    \Xbft(\omega)
    \end{matrix}\right] \PL2 \left[\begin{matrix}
        \sqrt{g(U)}Z_0\\ \alpha(\sqrt{g(U)}Z_0 + \tau(g(U)) Z_1)
        \end{matrix}\right],
\end{equation}
where $U\sim \unif([0, 2\pi])$, and $Z_0, Z_1 \sim\mcC\mcN(0,1)$, and $\tau(\cdot)$ is the function in \eqref{eq:tau_for_ridge}. Note that the variance of $\Xbft_0(\omega)$ is $g(\omega)$, hence the term $\tau(g(U))Z_1$. As mentioned earlier, this variance is the only variable that changes with frequency while all the other parameters are the same for all frequencies. Using this convergence, in the limit, the error is
\begin{multline}
    \lim_{n_x\rightarrow\infty}\lim_{T\rightarrow\infty}\frac{1}{n_x T}\fnormsm{\Xbfthat_\ridge-\Xbft}^2 \\= \int_0^{2\pi} \left((\alpha - 1)^2 g(\omega) + \alpha^2 \tau(g(\omega))\right)d\omega.\label{eq:proof_error_in_freq}
\end{multline}
As mentioned in Section \ref{sec:main_technical_lemmas}, our scaling of the Fourier operator makes it a unitary operator. Therefore, $\ell_2$ norm is preserved under our definition of Fourier transform and its inverse. This implies that the same result as in \eqref{eq:proof_error_in_freq} also holds in \emph{time} domain.

\section{Experiments}
In this section we validate our theoretical results on simulated data. We generate data using a ground truth convolutional model of the form \eqref{eq:conv_def}. We use i.i.d. complex normal convolution kernel and noise with different variances. For the data matrix $\Xbf$, we consider two different models: i) i.i.d. complex normal data; and ii) a non-Gaussian autoregressive process of order 1 (an AR(1) process). In both cases we take $T=256$, $n_y=500$ and use different values of $n_x$ to create plots of estimation error with respect to $\delta=n_y/n_x$.

AR(1) process is a process that evolves in time as
\begin{equation}
    x_{t} = {a} x_{t-1} + \xi_t, \label{eq:AR1}
\end{equation}
where $\xi_t$ is some zero mean random noise and ${a}$ is a fixed constant. Note that our assumptions on the process require it to be a stationary and ergodic process. These assumptions are only satisfied when the noise is i.i.d. and $|{a}|<1$. The parameter ${a}$ controls how fast the process is mixing. The case were ${a}=0$ results in an i.i.d. process, whereas values with magnitude close to 1 would result in a process that has strong correlations over a long period of time. This process has the form of one of the examples given in Section \ref{sec:random_process_assumptions} and hence satisfies all the assumptions required for our theory to hold. 

If the noise $\xi_t$ has zero mean Gaussian distribution, the process will be a centered Gaussian process which is completely characterized by an auto-correlation function
\begin{equation}
    R[t] = \Exp[x_{t'+t}x_{t'}],
\end{equation}
where we have used the fact that stationarity of the process implies this auto-correlation only depends on the time difference and not on the actual time. Since for Gaussian processes, the Fourier transform is another Gaussian process, we do not need to use the results of \cite{peligrad2010central} to analyze them. As such, a more interesting example would be to use a non-Gaussian noise $\xi$. Therefore, besides the Gaussian AR process, we also take $\xi_t\sim\unif(\{-s,s\})$ where $s$ is an step size that controls the variance of the process. The variance and auto-correlation of a univariate AR(1) process can be found as follows. Squaring both sides of \eqref{eq:AR1} and taking the expectation we obtain
\begin{equation}
    \Exp[x^2] = \frac{\Exp[\xi^2]}{1- {a}^2}.
\end{equation}
Similarly, it is also easy to show that the auto-correlation of this process is
\begin{equation}
    R[t] = \frac{\Exp[\xi^2]}{1- {a}^2}{a}^{|t|}.\label{eq:autocorrelation_AR1}
\end{equation}
The auto-correlation function of the AR process only depends on the variance of the noise and not its distribution. Our main result, Theorem \ref{thm:main}, shows that the asymptotic error of ridge estimator depends on the the underlying process only through the function $g(\omega)$ which as stated earlier, is proportional to the spectral density of the process, i.e. norm of Fourier transform of the auto-correlation function. Therefore, so long as the zero mean noise has the same variance, irrespective of its distribution, we expect to see the same asymptotic error in the convolutional ridge regression when the rows of $\Xbf$ are i.i.d. samples of such processes. To show this, we use both a Gaussian AR(1) process with $\var(\xi_t^2)=0.1$ as well as $\xi_t\unif(\{-s,s\})$ with $s=\sqrt{0.1}$ to match the variances. In both cases we take ${a}=0.9$ and measurement noise variance $\sigma^2 = 0.1$.
\begin{figure}[t]
    \centering
    \includegraphics[width=0.45\textwidth]{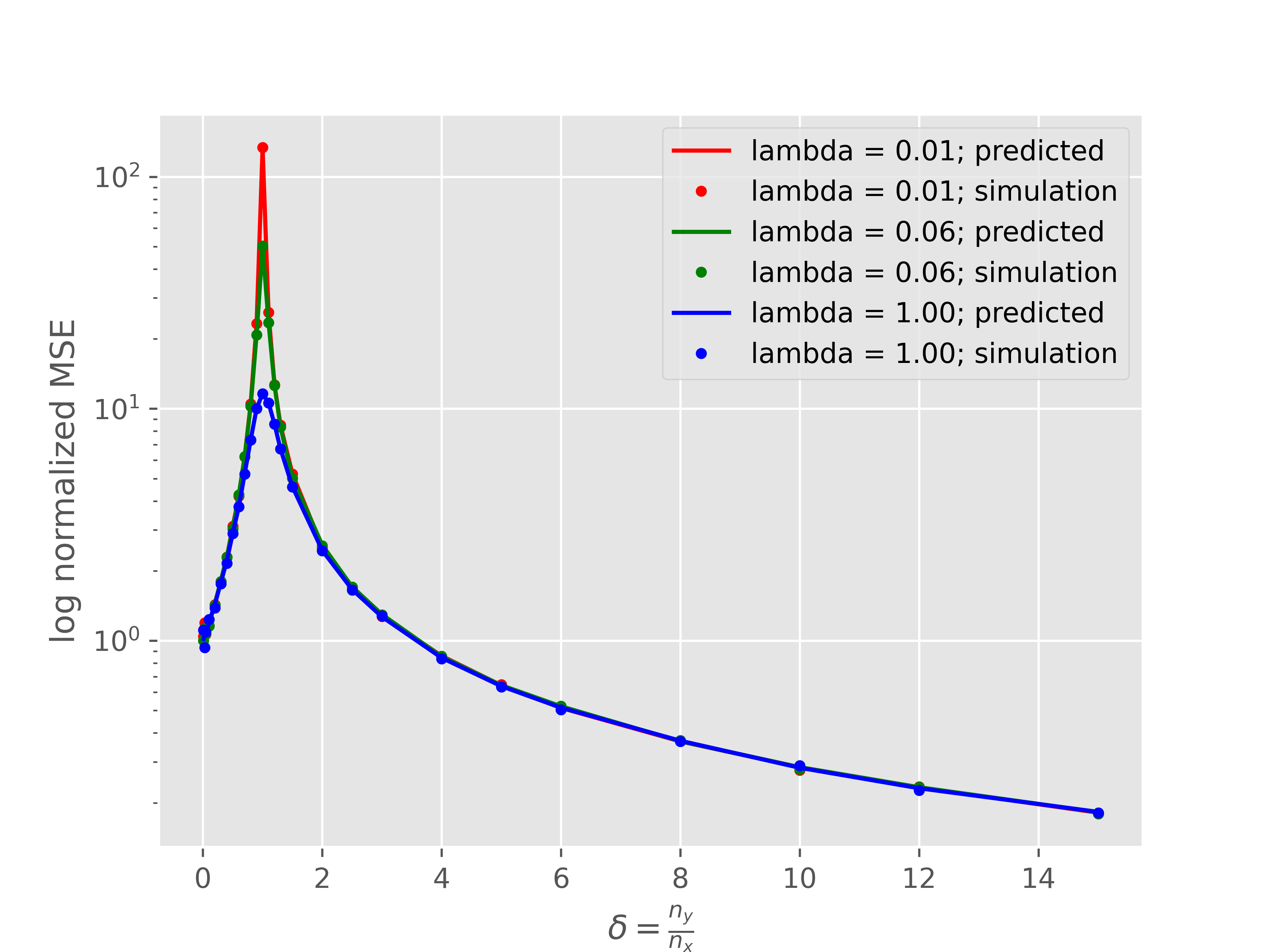}
    \caption{Log of normalized error for i.i.d. Gaussian features with respect to $\delta=n_y/n_x$ for three different values of $\lambda$. Solid lines show the predictions of our theory whereas the dots show the observed error on synthetic data. }
    \label{fig:iid_Gaussian}
\end{figure}
\begin{figure}[t]
    \centering
    \includegraphics[width=0.45\textwidth]{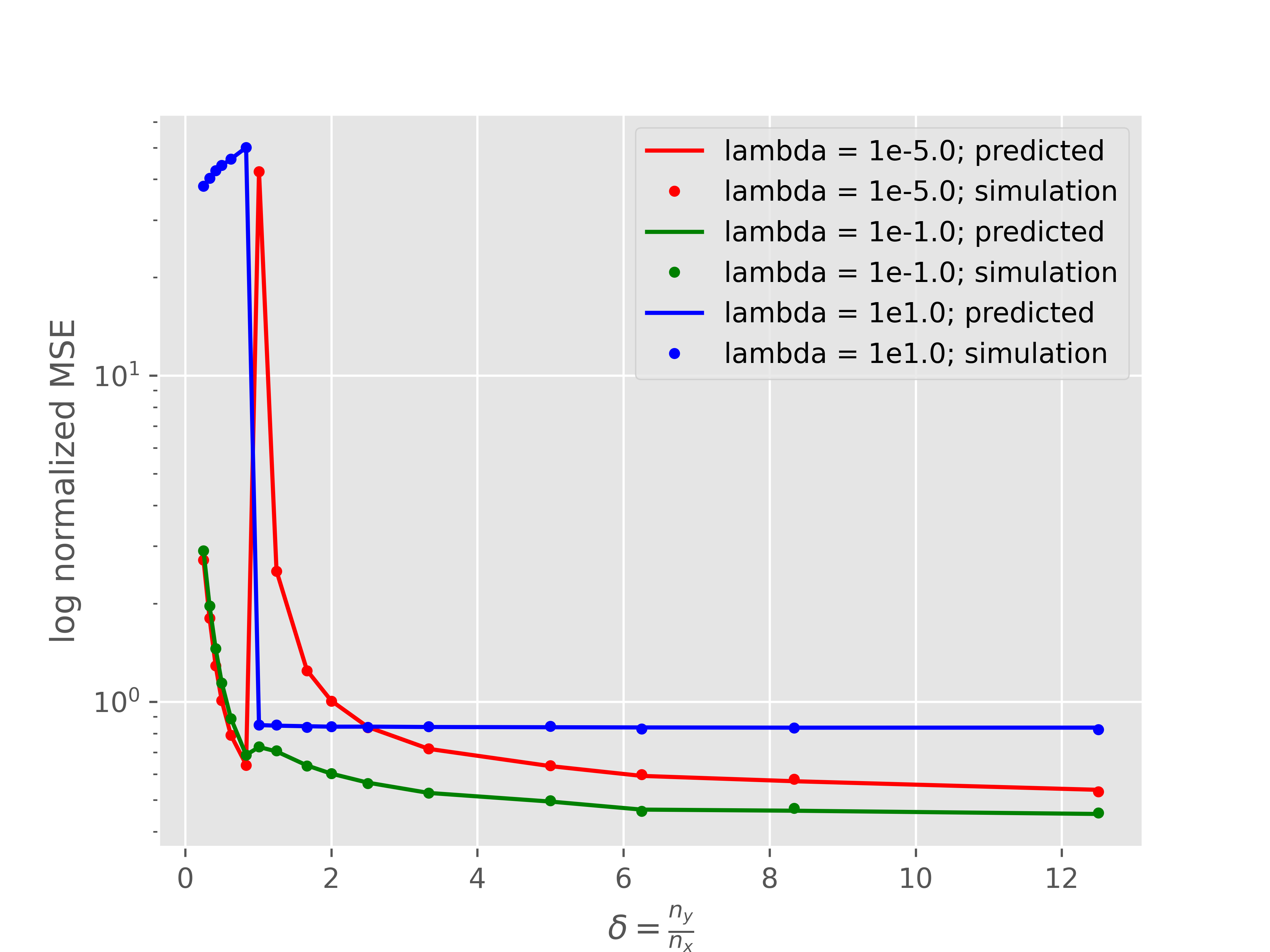}
    \caption{Log of normalized error for the AR(1) features with the process noise $\unif(\{-s, s\})$, with respect to $\delta=n_y/n_x$ for three different values of $\lambda$. Solid lines show the predictions of our theory and the dots show the observed error on synthetic data. The plots for Gaussian AR process is essentially identical. to this plot.}
    \label{fig:AR_random_walk}
\end{figure}

We first present the results for the i.i.d. Gaussian covariates. In this case the variance of signal and noise are $0.004$ and $1$ respectively. Figure \ref{fig:iid_Gaussian} shows the log of normalized estimation error with respect to $\delta=n_y/n_x$ for three different values of the regularization parameter $\lambda$. Normalized estimation error is defined as
\begin{equation}
    {\rm NMSE} = \frac{\Exp\|\Xbfhat_\ridge - \Xbf_0\|_2^2}{\Exp\tnorm{\Xbf_0}^2}.
\end{equation}
The solid curves correspond to what our theory predicts and the dots correspond to what we observe on synthetic data. Even though our results hold in the limit of $n_x, n_y, k, T\rightarrow \infty$ at proportional ratio, we can see that already at this problem size, there is an almost perfect match between our predictions and the error that is observed in practice. This suggests that the errors concentrate around these asymptotic values. The figure also shows the double descent phenomenon where as the number of parameters increases beyond the interpolation threshold, the error starts decreasing again. It can be seen that regularization helps with pulling the estimation error down in vicinity of the interpolation threshold. The interpolation threshold is where we have just enough parameters to fit the observations perfectly. This happens at $\delta=1$, i.e.\ $n_x = n_y$.

As mentioned in Remark \ref{rem:double_descent_iid} an i.i.d. process has a white spectrum in frequency domain, meaning that $\g(\omega)$ is a constant. Therefore, for these processes, the integral in Corollary \ref{cor:estimation_error} would be proportional to the integrand. The integrand in turn is the asymptotic error of an ordinary ridge regression problem with i.i.d. Gaussian features. As such, this figure is essentially the same as the figures in papers that have looked at asymptotics of ridge regression, some of which we have mentioned in the Introduction and prior work.

Figure \ref{fig:AR_random_walk} shows the same plot for the case where the rows of $\Xbf$ are i.i.d. samples of an AR(1) process with process noise $\unif(\{-s, s\})$. The asymptotic error for processes that have dependencies over time can be significantly different from i.i.d. random features. The red curve is similar to the red curve in Figure \ref{fig:iid_Gaussian}, but the other curves show very different behavior. The double descent phenomenon is still present here. The plot for AR(1) process with Gaussian noise is essentially identical to Figure \ref{fig:AR_random_walk} and we have moved it to the appendix (Figure \ref{fig:AR_Gaussian}). This supports our theoretical result that error in this asymptotic regime only depends on the spectral density of the process.

\section{Conclusion}
\paragraph{Summary.}We characterized the performance of ridge estimator for convolutional models in proportional asymptotics regime. By looking at the problem in Fourier domain, we showed that the asymptotic mean squared error of ridge estimator can be found from a scalar integral that depends on the spectral properties of the true signal. Our experiments show that our theoretical predictions match what we observe in practice even for problems of moderate size.

\paragraph{Future work.} The results of this work only apply to ridge regression estimator for convolutional linear inverse problem. The key property of ridge regularization is that it is invariant under unitary transforms, and hence we could instead solve the problem in frequency domain. Proving such result for general estimators and regularizers allows us to extend this work to inference in deep convolutional neural networks similar to \cite{pandit2020inference}. Such work would allow us to obtain the estimation error inverse problems use of deep convolutional generative priors.

\bibliography{ref}
\newpage
\appendix

\clearpage
\appendix
\section{Complex Normal Distribution}\label{app:complex_normal_dist}
Complex normal is the distribution of a complex random variable whose imaginary and real parts are jointly Gaussian.

\paragraph{Standard complex normal distribution.} A random variable $Z = X + \iu Y$ where $X,Y\in \Real$ has standard complex normal distribution represented by $\CNorm(0,1)$ if
\[X, Y\sim \Norm(0,1/2), \quad X\indep Y.\]

\paragraph{General complex Gaussian distribution.} A random vector $\Zbf = \Xbf + \iu \Ybf$ where $\Xbf, \Ybf\in \Real^n$ has complex Gaussian distribution $\CNorm(\mubf, \Gammabf, \Cbf)$ if $\Xbf$ and $\Ybf$ are jointly Gaussian with
\begin{align}
    \mubf &= \Exp[\Zbf],\\
    \Gammabf &= \Exp[(\Zbf-\mubf)(\Zbf -\mubf)\herm],\\
    \Cbf &= \Exp[(\Zbf-\mubf)(\Zbf -\mubf)\tran].
\end{align}
The parameters $\mubf, \Gammabf$, and  $\Cbf$ are called mean vector, covariance matrix, and relation matrix respectively. Alternatively, if we define
\begin{align*}
    \Cbf_{XX} &= \Exp[(\Xbf-\mubf_X)(\Xbf -\mubf_x)\tran], \quad \mubf_X = \Exp[\Xbf],\\
    \Cbf_{YY} &= \Exp[(\Ybf-\mubf_Y)(\Ybf -\mubf_Y)\tran], \quad \mubf_X = \Exp[\Ybf],\\
    \Cbf_{XY} &= \Cbf_{YX}\tran = \Exp[(\Xbf-\mubf_X)(\Ybf -\mubf_Y)\tran],
\end{align*}
then $\Xbf, \Ybf$ are jointly Gaussian with distribution
\begin{equation*}
    (\Xbf, \Ybf)\sim \Norm\left(\left[\begin{matrix}\mubf_X\\\mubf_Y
    \end{matrix}\right], \left[\begin{matrix}\Cbf_{XX} & \Cbf_{XY}\\\Cbf_{YX} & \Cbf{YY}
    \end{matrix}\right]\right).
\end{equation*}
The matrices $\Gammabf$ and $\Cbf$ are related to covariance matrices of $\Xbf$ and $\Ybf$ through the following equations:
\begin{align*}
    \Gammabf &= \Cbf_{XX} + \Cbf_{YY} + \iu (\Cbf_{YX} - \Cbf_{XY}),\\
    \Gammabf &= \Cbf_{XX} - \Cbf_{YY} + \iu (\Cbf_{YX} + \Cbf_{XY}).
\end{align*}

\section{Empirical Convergence of Vector Sequences}\label{app:empirical_conv_definitions}
Here we review some definitions that are standard to in papers that use approximate message passing.

\begin{definition}[Pseudo Lipschitz Continuity] A function $\fbf$ is called pseudo-Lipschitz continuous of order $p$ with constant $C$ if for all $ \xbf_1, \xbf_2\in \dom(\fbf)$
\begin{equation}
    \norm{\fbf(\xbf_1)-\fbf(\xbf_2)}\leq C\norm{\xbf_1-\xbf_2}(1+\norm{\xbf_1}^{p-1} + \norm{\xbf_2}^{p-1}).
    \end{equation}
\end{definition}
Note that for $p=1$ this definition is a equivalent to the definition of the standard Lipschitz-continuity.

\begin{definition}[Uniform Lipschitz-continuity] A function $\fbf$ on $\mcX\times \mcW$ is \emph{uniformly Lipschitz-continuous} in $\xbf$ at $\bar{\omegabf}$ if there exists constants $L_1, L_2\geq 0$ and an open neighborhood $U$ of $\bar{\omegabf}$ such that for all $\xbf_1,\xbf_2 \in \mcX, \omegabf \in U$
\begin{equation}
    \norm{\fbf(\xbf_1, \omegabf) - \fbf(\xbf_2, \omegabf)} \leq L_1 \norm{\xbf_1-\xbf_2},
\end{equation}
and for all $\xbf\in \mcX, \omegabf_1, \omegabf_2\in \mcW$
\begin{equation}
    \norm{\fbf(\xbf, \omegabf_1)-\fbf(\xbf, \omegabf_2)} \leq L_2(1+ \norm{\xbf})\norm{\omegabf_1-\omegabf_2}.
\end{equation}
\begin{definition}[Empirical convergence of sequences] Consider a sequence of vectors $\xbf(N) = \{\xbf_n(N)\}_{n=1}^N$ with $\xbf_n(N)\in \Real^d$, i.e.\ each $\xbf(N)$ is a block vector with a total of $Nd$ components. For a finite $p\geq 1$, we say that the vector sequence $\xbf(N)$ converges empirically with $p$th order moments if there exists a random variable $X\in \Real^d$ such that

\begin{itemize}
    \item $\Exp\norm{X}_p^p <\infty$;
    \item for any $f:\Real^d\rightarrow \Real$ that is pseudo-Lipschitz of order $p$,
    \begin{equation}
        \lim_{N\rightarrow \infty}\frac{1}{N}\sum_{n=1}^N f(\xbf_n(N))=\Exp[f(X)].\label{eq:PL2_function_convergence}
    \end{equation}
    
\end{itemize}
\end{definition}
With some abuse of notation, we represent this with
\begin{equation}
    \lim_{N\rightarrow \infty}\xbf_n \PLp X,
\end{equation}
where we have omitted the dependence on $N$ to ease the notation. In this definition the sequence $\{\xbf(N)\}$ can be random or deterministic. If it is random we require the equality in \eqref{eq:PL2_function_convergence} to hold almost surely. In particular, if the sequence $\{\xbf_n\}$ is i.i.d.\ with $\xbf_n\sim p_X(\cdot)$, with $\Exp\norm{X}_p^p <\infty$, then $\{\xbf_n\}$ converges empirically to $X$ with $p$th order. The extension of this definition to sequence of matrices and higher order tensors is straightforward. 
\begin{definition}[Convergence in distribution]
A sequence of random vectors $\xbf_n\in\Real^d$ converges in distribution (also known as weak convergence) to $\xbf$ if for all bounded functions $f:\Real^d\rightarrow \Real$
\begin{equation}
    \lim_{n\rightarrow \infty}\Exp f(\xbf_n) = \Exp f(\xbf).
\end{equation}
\end{definition}
PL(p) convergence is equivalent to convergence in distribution plus convergence of the $p$th moment \cite{bayati2011dynamics}.
\begin{definition}[Wasserstein-$p$ distance]
Wasserstein-$p$ distance between two probability measures $\mu, \nu$ on Euclidean space $\Real^d$ is
\begin{equation}
    W_p(\mu, \nu) = \inf_{\gamma\in \Gamma} \left(\Exp_{(\xbf, \ybf)\sim \gamma}\norm{\xbf-\ybf}_p^p\right)^{\frac{1}{p}},
\end{equation}
where $\Gamma$ is the set of all probability measures on the product space $\Real^d\times \Real^d$ with marginals $\mu$ and $\nu$. 
\end{definition}
PL(p) convergence is also equivalent to convergence the empirical measure of the sequence $\xbf_n$ to probability measure of $X$ in Wasserstein-$p$ distance \cite{villani2008optimal}.  
\end{definition}

\section{1D Convolution Operators in Matrix Form}
In this section we derive the matrix form of 1D convolution operators to show how these operators look like in time domain. As we will see, convolution operators in time domain can be represented as a \emph{doubly block circulant matrix}. Because of this structure, approximate message passing (AMP) (discussed in Appendix \ref{app:AMP}) cannot be directly used to obtain estimation error of ridge regression for convolutional inverse problem in time domain. This is due to the assumption in AMP that the measurement matrix has i.i.d.\ entries. If this assumption can be relaxed, we can analyze estimators other than ridge, and compute error metrics other than MSE. We hope to follow this direction in a future work.

First assume that in the convolutional model in \eqref{eq:conv_model}, $n_x=n_y=1$, i.e. the input and output both have one channel. Also for a matrix $\Zbf\in \Real^{m\times n}$, let $\vec(\Zbf)\in \Real^{nm}$ represent the vector constructed by stacking $\Xbf$ in a vector row by row. To simplify the notation, we zero pad the convolution kernel which in this case is a vector of size $k$, so that it will have size $T$ and we still use $\Kbf$ to represent the zero-padded kernel to simplify the notation. In this case, the convolution operator $\Kbf:\Xbf\mapsto \Kbf*\Xbf$ can be represented as a circulant matrix $\Cbf: \vect(\Xbf)\mapsto \Cbf \vect(\Xbf)$
\begin{equation}\label{eq:conv_1channel_matrix_form}
    \Cbf = \left[
    \begin{matrix}
    K_1 & K_2 & K_3 & \dots &  K_T\\
    K_T & K_1 & K_2 & \dots &K_{T-1}\\
    K_{T-1} & K_{T} & K_1 & \dots & K_{T-3}\\
    \vdots & \vdots & \vdots & \ddots & \vdots\\
    K_2 & K_3 & K_4 & \dots & K_1
    \end{matrix}
    \right]
\end{equation}
When the number of input channels and output channels are $n_x$ and $n_y$ respectively, the convolution can be represented in matrix form as matrix with blocks of circulant matrices
\begin{equation}
    \Cbf = \left[
    \begin{matrix}
    \Cbf_{11} & \Cbf_{12}  & \dots &  \Cbf{1,n_x}\\
    \Cbf_{21} & \Cbf_{22}  & \dots &  \Cbf{2,n_x}\\
    \vdots & \vdots & \ddots & \vdots\\
    \Cbf_{n_y,1} & \Cbf_{n_y,2}  & \dots &  \Cbf{n_y,n_x}
    \end{matrix}
    \right],\label{eq:conv_matrix_form}
\end{equation}
where each $\Cbf_{ij}$ is a circulant matrix of the form \eqref{eq:conv_1channel_matrix_form} constructed from $\Kbf_{ij*}$. Since the adjoint of a circulant matrix is also a circulant matrix, one can see that the adjoint of a 1D convolution (with stride 1) is also a convolution with respect to another kernel.

\section{Approximate Message Passing}\label{app:AMP}
In this section we briefly describe the approximate message passing (AMP) algorithm for linear inverse problems \cite{bayati2011dynamics}. Consider the problem of estimating $\xbf^0$ from linear observations
\begin{equation}
    \ybf = \Abf \xbf^0 + \xibf,\label{eq:linear_inverse_prob}
\end{equation}
where $\Abf \in\Real^{n_y\times n_x}$ is a known matrix and $\xibf$ is i.i.d.\ zero-mean Gaussian noise with variance $\sigma^2$. Approximate message passing is an iterative algorithm to solve this problem
\begin{align}
    \xbf^{t+1} &= \etabf_t(\Abf\tran \zbf^t + \xbf^t)\label{eq:AMP_x_update_general}\\
    \zbf^t &= \ybf - \Abf \xbf^t + \underbrace {\frac{1}{\delta}\zbf^{t-1}\inner{\etabf'_{t-1}(\Abf\tran \zbf^{t-1} + \xbf^{t-1})}}_{\text{Onsager correction}},\label{eq:AMP_z_update_general}
\end{align}
where $\etabf_t(\cdot)$ is a denoiser that acts component-wise, and $\inner{\cdot}$ is the empirical averaging operator.

The key property of AMP algorithm is that when the sensing matrix $\Abf$ is large with i.i.d. sub-Gaussian entries, the behavior of the algorithm at each iteration can be exactly characterized via a \emph{scalar} recursive equation called the \emph{state evolution} (SE) 
\begin{equation}
    \tau_{t+1}^2 = \sigma^2 +\frac{1}{\delta}\Exp\left[(\eta_t(X_0 + \tau_t Z)-X_0)^2\right],\label{eq:AMP_SE}
\end{equation}
where $X_0\sim p_{X_0}$ independent of $Z\sim \Norm(0,1)$. Here $p_{X_0}$ is the distribution to which the components of $\xbf^0$ are converging empirically. See Appendix \ref{app:empirical_conv_definitions} for background on empirical convergence of sequences and some definitions we would use throughout this paper. Given $\tau_t$, as $n_x, n_y\rightarrow \infty$ with fixed ratio $\delta :=n_y/n_x$ we have
\begin{equation}
    \left[\begin{matrix} \xbf_0\\
    \xbf^{t}
    \end{matrix}\right] \PL2 \left[\begin{matrix} X_0\\
    \eta_{t-1}(X_0 + \tau_{t-1} Z)
    \end{matrix}\right],\label{eq:AMP_PL2convergence}
\end{equation}
where as in SE we have $X_0\sim p_{X_0}$ independent of $Z\sim \Norm(0,1)$.

This convergence allows us to compute the estimation error. If we define the mean squared error of the estimate at iteration $t$ to be ${\rm MSE}=1/n_x\tnorm{\xbf^0 - \xbf^t}^2$, then in the large system limit almost surely 
\begin{equation}
    {\rm MSE} = \Exp\left[\left(\etabf_{t-1}(X_0 + \tau_{t-1} Z)-X_0\right)^2\right],\label{eq:MSE_AMP}
\end{equation}
where the expectation is over $X_0$ and $Z$.

\subsection{AMP for ridge regression}\label{app:AMP_ridge}
In this section we show how AMP can be used to derive asymptotic error of ridge regression
\begin{equation}
    \xbfhat_\ridge = \argmin_\xbf \tnorm{\ybf - \Abf\xbf}^2 + \lambda \tnorm{\xbf}^2.
\end{equation}
The solution to this optimization problem is
\begin{equation}
    \xbfhat_\ridge = (\Abf\tran \Abf + \lambda \Ibf)\inv \Abf\tran \ybf.\label{eq:ridge_closed_form_Sol}
\end{equation}
Next, consider the AMP recursion in \eqref{eq:AMP_x_update_general} and \eqref{eq:AMP_z_update_general} with a fixed denoiser $\etabf_t(\xbf) = \alpha \xbf$
\begin{align}
    \xbf^{t+1} &= \alpha(\Abf\tran \zbf^t + \xbf^t),\label{eq:AMP_ridge_x_update}\\
    \zbf^t &= \ybf - \Abf \xbf^t + \frac{\alpha}{\delta}\zbf^{t-1}.\label{eq:AMP_ridge_z_update}
\end{align}
The next lemma shows that this recursion solves the ridge regression for a specific regularization parameter $lambda$.
\begin{lemma}\label{lem:lambda_corresponding_to_alpha}
The fixed point of AMP algorithm with $\etabf_t(\xbf) = \alpha \xbf$ is the solution of ridge regression with
\begin{equation}
    \lambda = \frac{(1-\alpha)(1- \alpha/\delta)}{\alpha},\label{eq:lamda_in_terms_of_alpha}
\end{equation}
where $\delta=n_y/n_x$.
\end{lemma}
\begin{proof}
Let $\xbf^*$ and $\ybf^*$ denote the fixed points of the AMP recursion. Then we have
\begin{align}
    \xbf^* &= \alpha(\Abf\tran \zbf^* + \xbf^*),\label{eq:AMP_ridge_fixed_point1}\\
    \zbf^* &= \ybf - \Abf \xbf^* + \frac{\alpha}{\delta}\zbf^*.
\end{align}
Therefore,
\begin{equation}
    \zbf^* = \frac{1}{1 - \alpha/\delta}(\ybf - \Abf\xbf).
\end{equation}
Plugging this back to Equation \eqref{eq:AMP_ridge_fixed_point1} we get
\begin{equation}
    \xbf^* = \left(\Abf\tran\Abf + \frac{(1-\alpha)(1-\alpha/\delta)}{\alpha}\Ibf\right)\inv\Abf\tran\ybf.
\end{equation}
Comparing this to \eqref{eq:ridge_closed_form_Sol} proves the result.
\end{proof}
Given a $\lambda$, one can solve the quadratic equation \eqref{eq:lamda_in_terms_of_alpha} to find the $\alpha$ that satisfies the equation. This is a quadratic equation that has two solutions. As we show in Section \ref{app:stability_of_AMP}, so long as the regularization parameter $\lambda$ is non-negative, this quadratic equation always has two real and positive solutions. But only for the smaller solution the AMP recursions for solving ridge regression converges, and hence only the smaller one is valid. 

Having found the $\alpha$ we can use the state evolution \eqref{eq:AMP_SE} to find its fixed point. For ridge regression, this can be done in closed form. The state evolution for ridge regression can be written as
\begin{equation}
    \tau_{t+1}^2 = \sigma^2 + \frac{1}{\delta}((1-\alpha)^2\sigma_X^2 + \alpha^2 \tau_t^2),\label{eq:state_evolution_ridge}
\end{equation}
If we define the fixed point value $\tau := \lim_{t\rightarrow \infty}\tau_t$ we have that it should satisfy
\begin{equation}
    \tau^2 = \sigma^2 + \frac{1}{\delta}((1-\alpha)^2\sigma_X^2 + \alpha^2 \tau^2),
\end{equation}
from which we obtain
\begin{equation}
    \tau^2 = \frac{\sigma^2 +\frac{1}{\delta}(1-\alpha^2)\sigma_X^2}{1 - \frac{\alpha^2}{\delta}}.
\end{equation}
The mean squared error then can be obtained as
\begin{align*}
    \frac{1}{n_x}\tnorm{\xbfhat_\ridge- \xbf_0}^2 &=\Exp\left[\left(\alpha(X_0 + \tau Z)-X_0\right)^2\right]\\
    &= (\alpha - 1)^2\Exp X_0^2 + \alpha^2 \tau^2.
\end{align*}

\subsection{Convergence of AMP}\label{app:stability_of_AMP}
As mentioned in the previous section, when we use AMP to find the solution of ridge regression, we first need to find an $\alpha$ that satisfies Equation \eqref{eq:lamda_in_terms_of_alpha}. This is a quadratic equation that has two solutions. In theory, the solution of ridge regression with a given $\lambda$ is the fixed points of AMP iterations for both values of $\alpha$. However, we should also note that the results of AMP are only valid if the iterations converge to a fixed point. This is equivalent to stability of the dynamics corresponding to AMP recursion. We saw in Lemma \ref{lem:lambda_corresponding_to_alpha} that a linear denoiser $\eta_t(\xbf) = \alpha \xbf$ can be used to solve for a ridge regression with regularization parameter $\lambda$. Recall that the AMP iterations for this denoiser are
\begin{align}
    \xbf^{t+1} &= \alpha(\Abf\tran \zbf^t + \xbf^t)\label{eq:AMP_x_update}\\
    \zbf^t &= \ybf - \Abf \xbf^t + \frac{\alpha}{\delta}\zbf^{t-1}.\label{eq:AMP_z_update}
\end{align}
Plugging Equation \eqref{eq:AMP_z_update} in Equation \eqref{eq:AMP_x_update} we get
\begin{align}
    \xbf^{t+1} &= \alpha(\Ibf - \Abf\tran\Abf) \xbf^t +\frac{\alpha^2}{\delta}\zbf^{t-1} + \alpha\Abf\tran \ybf,\\
    \zbf^t &= \ybf - \Abf \xbf^t + \frac{\alpha}{\delta}\zbf^{t-1}.
\end{align}
These equations correspond to a linear time invariant system with state matrix
\begin{equation}
    \mcA = \left[\begin{matrix}\alpha(\Ibf - \Abf\tran \Abf) & \frac{\alpha^2}{\delta}\Abf\tran\\ -\Abf & \frac{\alpha}{\delta}\Ibf\end{matrix}\right].
\end{equation}
The system is stable if and only if all the eigenvalues of $\mcA$ lie inside the unit disk. A simple row operation (which does not change the eigenvalues) shows that the eigenvalues of $\mcA$ are the same as eigenvalues of
\begin{equation}
    \mcA' = \left[\begin{matrix}\alpha\Ibf  & \zero\\ -\Abf & \frac{\alpha}{\delta}\Ibf\end{matrix}\right].
\end{equation}
Therefore, the AMP recursions in \eqref{eq:AMP_x_update} and \eqref{eq:AMP_z_update} are stable, i.e.\ converge to the fixed points corresponding to the ridge regression if and only if
\begin{equation}
    | \alpha| \leq 1, \quad | \frac{\alpha}{\delta}|\leq 1.\label{eq:stability_alpha_ineqs}
\end{equation}
Since $\delta> 0$, this is equivalent to
\begin{equation}
    |\alpha| \leq \min(1, \delta).\label{eq:stability_condition_alpha}
\end{equation}
If regularization parameter $\lambda \geq 0$, solving the quadratic equation \eqref{eq:lamda_in_terms_of_alpha} for $\alpha$, it is not hard to show that it has two solutions $\alpha_1, \alpha_2$ that are always real and satisfy
\begin{equation}
    0< \alpha_1 \leq \min(1, \delta) \leq \max(1, \delta) \leq \alpha_2.
\end{equation}
Comparing this to \eqref{eq:stability_condition_alpha}, we see that only $\alpha_1$ satisfies the stability condition. To summarize, \eqref{eq:lamda_in_terms_of_alpha} always has two real positive solutions, but only the smaller one satisfies the stability condition. 

As a sanity check, we can also verify that if AMP iterations for ridge regression in  \eqref{eq:AMP_x_update} and \eqref{eq:AMP_z_update} are stable, so is the state evolution recursion. The state evolution for ridge regression is given in \eqref{eq:state_evolution_ridge}. This is a scalar linear time invariant system that is stable if and only if 
\begin{equation}
    -1\leq \frac{\alpha^2}{\delta}\leq 1.
\end{equation}
Clearly, the stability conditions in \eqref{eq:stability_alpha_ineqs} imply this inequality. Therefore, the stability of AMP recursions for ridge regression also implies the stability of the state evolution for ridge regression. As a result, the smaller value of $\alpha$ that satisfies \eqref{eq:lamda_in_terms_of_alpha} should be used to get the correct prediction of error.

\subsection{AMP for complex ridge regression}\label{app:complex_AMP}
Approximate message passing can also be used when the signals in \eqref{eq:linear_inverse_prob} are complex valued. So long as the sensing matrix has i.i.d. complex normal entries $\Abf_{ij}\sim 
\CNorm(0, \sigma_A^2/n_y)$ (see Appendix \ref{app:complex_normal_dist} for a brief overview of complex normal distribution), i.e.\ the real and imaginary parts of each entry are i.i.d. Gaussian random variables with variance $\sigma_A^2/(2n_y)$ and independent of each other, the state evolution holds \cite{maleki2013asymptotic}. Therefore, by changing all variables to complex variables, we can use AMP exactly as in Appendix \ref{app:AMP_ridge} and get the asymptotic error of complex ridge regression using the state evolution almost without any changes.

\section{Experiment with Gaussian AR(1) Process}
\begin{figure}[t]
    \centering
    \includegraphics[width=0.5\textwidth]{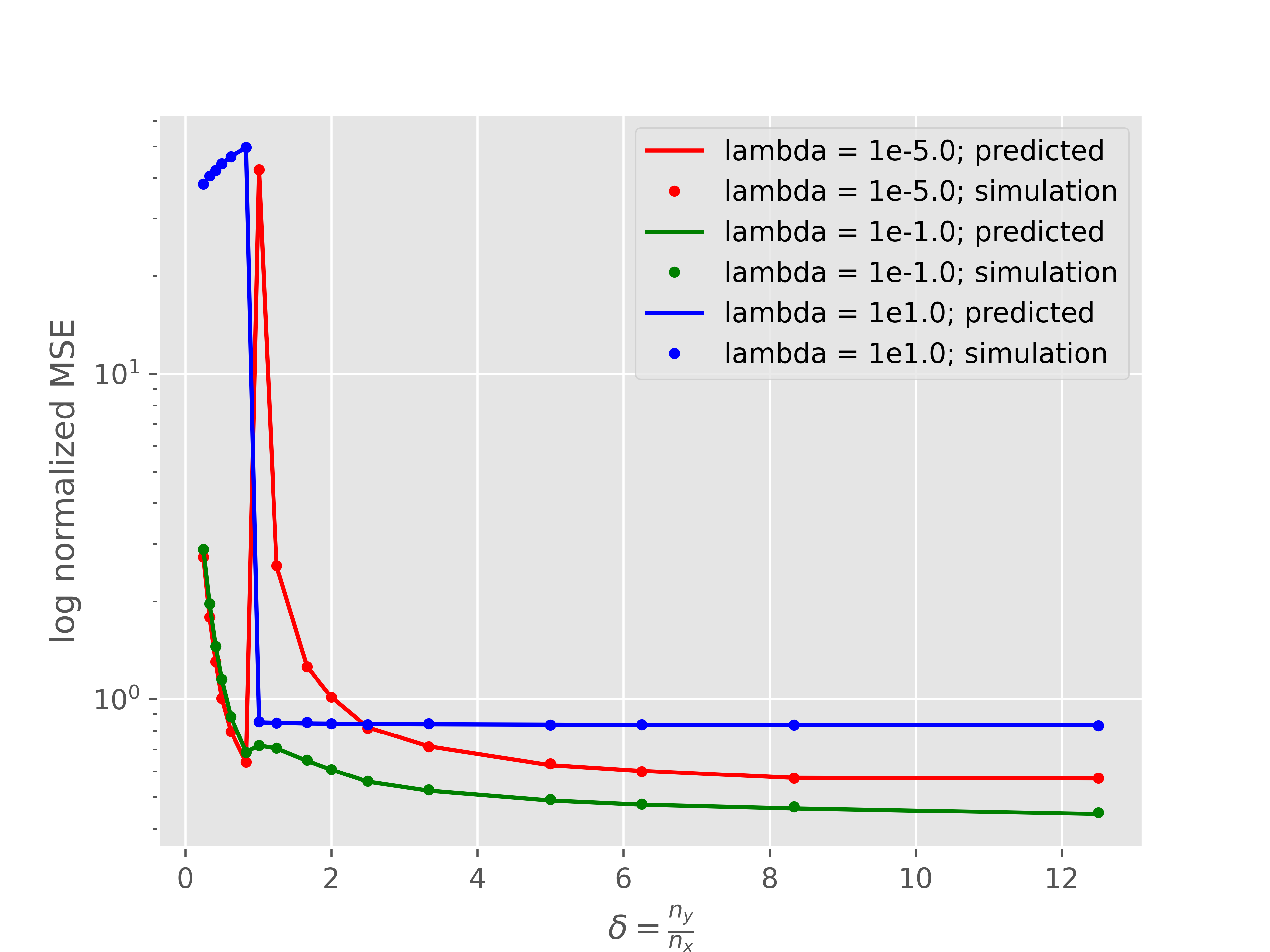}
    \caption{Log of normalized error for the AR(1) features with the process noise $\Norm(0, s^2)$, with respect to $\delta=n_y/n_x$ for three different values of $\lambda$. The figure is almost indistinguishable from Figure \ref{fig:AR_random_walk}.}
    \label{fig:AR_Gaussian}
\end{figure}
As mentioned in the experiments, for an AR(1) process as in \eqref{eq:AR1}, the auto-correlation function derived in Equation \eqref{eq:autocorrelation_AR1} does not depend on the distribution of the noise $\xi_t$, but only its second moment. This is true in general for an AR($p$) process that evolves as a linear time-invariant (LTI) system driven with zero-mean i.i.d. noise. For such processes the auto-correlation only depends on the second order statistics of the noise as well parameters of the linear system. Therefore, we expect to get identical results in the limit if the any zero mean noise is driving the process so long as the variances match. In Figure \ref{fig:AR_random_walk}, we showed the results for the case where the noise was a scaled Rademacher random variable. Figure \ref{fig:AR_Gaussian} shows the same results for the case where the noise is Gaussian with the matched variance. As expected, this plot is almost indistinguishable from Figure \ref{fig:AR_random_walk}.

\end{document}